\documentclass{article}

\pdfoutput=1

\PassOptionsToPackage{numbers}{natbib}

    \usepackage[preprint]{neurips_2023}

\usepackage[utf8]{inputenc} 
\usepackage[T1]{fontenc}    
\usepackage{hyperref}       
\usepackage{url}            
\usepackage{booktabs}       
\usepackage{amsfonts}       
\usepackage{nicefrac}       
\usepackage{microtype}      
\usepackage{xcolor}         

\usepackage{amsmath}
\usepackage{amssymb}
\usepackage{amsthm}
\usepackage{graphicx}
\usepackage{multirow}
\usepackage{caption}
\usepackage{wrapfig}

\usepackage[capitalize]{cleveref}

\newtheorem{theorem}{Theorem}

\newenvironment{proofidx}[1]{%
  \renewcommand{\proofname}{Proof of Theorem #1}\proof}{\endproof}

\title{Deep Equilibrium Multimodal Fusion}

\author{%
    Jinhong Ni \\
    \And
    Yalong Bai \\
    \And
    Wei Zhang \\
    \And
    Ting Yao \\
    \And
    Tao Mei \\
}

\begin{document}

\maketitle

\begin{abstract}
  Multimodal fusion integrates the complementary information present in multiple modalities and has gained much attention recently. Most existing fusion approaches either learn a fixed fusion strategy during training and inference, or are only capable of fusing the information to a certain extent. Such solutions may fail to fully capture the dynamics of interactions across modalities especially when there are complex intra- and inter-modality correlations to be considered for informative multimodal fusion. In this paper, we propose a novel deep equilibrium (DEQ) method towards multimodal fusion via seeking a fixed point of the dynamic multimodal fusion process and modeling the feature correlations in an adaptive and recursive manner. This new way encodes the rich information within and across modalities thoroughly from low level to high level for efficacious downstream multimodal learning and is readily pluggable to various multimodal frameworks. Extensive experiments on BRCA, MM-IMDB, CMU-MOSI, SUN RGB-D, and VQA-v2 demonstrate the superiority of our DEQ fusion. More remarkably, DEQ fusion consistently achieves state-of-the-art performance on multiple multimodal benchmarks. The code will be released.
\end{abstract}

\section{Introduction}
Humans routinely receive and process signals through interactions across multiple modalities, supporting the unique human capacity to perceive the world. With the rise and development of deep learning, there has been a steady momentum of innovation that leverage multimodal data for learning deep models \cite{ngiam2011multimodal,mroueh2015deep,ramachandram2017deep}. Multimodal fusion, the essence of multimodal learning, aims to integrate the information from different modalities into a unified representation, and has made great success in real-world applications, \emph{e.g.}, sentiment analysis \cite{zadeh2016mosi}, multimodal classification \cite{arevalo2017gated}, medical analysis \cite{banos2015design,wang2021mogonet}, object detection~\cite{song2015sun}, visual question answering~\cite{goyal2017making}, \textit{etc}.

A common practice for deep multimodal learning is to first exploit modality-specific deep neural networks to extract modality-wise features, and then capitalize on multimodal fusion to combine the information from all modalities. The recent progress in computer vision and natural language processing area has convincingly pushed the limits of modality-specific learning \cite{he2016deep,vaswani2017attention,liu2021swin}, whereas multimodal fusion remains challenging for multimodal learning. Most conventional approaches are dedicated to deliberately designing fusion strategies \cite{liu2018efficient,ortega2019multimodal,nagrani2021attention}, which have proceeded along three dimensions of early fusion, mid fusion, and late fusion, with respect to the placement of fusion module in the whole framework. In general, these fusion strategies perform \emph{statically} during training and inference, \textit{i.e.}, the fusion architectures are often fixed. As a result, these approaches seldom explore modality importance, and the exchange of information within and across modalities is reinforced only to a certain degree. That might result in the generalization problem to various multimodal tasks, especially for some complicated multimodal correlations, \textit{e.g.}, the evolving temporal modality correlations. Moreover, for simple modality inputs, these static approaches might be excessive and potentially encode redundant, unstable, and even noisy information.

In an effort to improve the static fusion approaches, recent works endow the fusion mechanism with more power of leveraging three ways: 1) stabilizing and aligning signals from different modalities~\cite{duan2022multi}; 2) integrating interactions across modalities ranging from low level to high level~\cite{hou2019deep,pan2020x}; 3) dynamically perceiving the effective information and removing the redundancy from each modality~\cite{han2022multimodal,xue2022dynamic}. To the best of our knowledge, there is no unified multimodal fusion framework that looks into all three aspects simultaneously. This motivates us to develop a dynamic multimodal fusion architecture to adaptively model the cross-modality interactions from low level, middle level, to high level, making the architecture generic for various multimodal tasks.

To consolidate the above idea, we present a new deep equilibrium (DEQ) method for multimodal fusion in this paper. Our launching point is to recursively execute nonlinear projections on modality-wise features and the fused features until the equilibrium states are found. Specifically, our contributions include: 1) we seek the equilibrium state of features to jointly stabilize intra-modality representations and inter-modality interactions; 2) our method continuously applies nonlinear projections to modality-wise features and the fused features in a recursive manner. As such, the cross-modality interactions are reinforced at every level for multimodal fusion; 3) we devise a \emph{purified-then-combine} fusion mechanism by introducing a soft gating function to dynamically perceive modality-wise information and remove redundancy. Our DEQ fusion generalizes well to various multimodal tasks on different modalities and is readily pluggable to existing multimodal frameworks for further improvement. 

We evaluate our DEQ fusion approach on several multimodal benchmarks built on different modalities, including medical breast invasive carcinoma PAM50 subtype classification on BRCA, image-text movie genre classification on MM-IMDB,
audio-text sentiment analysis on CMU-MOSI, RBG-point 3D object detection on SUN RGB-D, and image-question visual question answering on VQA-v2. 
Our DEQ fusion approach consistently achieves new state-of-the-art performance on all benchmarks, demonstrating the superiority of modeling modality information from low level to high level in a dynamic way for multimodal fusion.

\section{Related Works}

\textbf{Multimodal Fusion} aims to integrate modality-wise features into a joint representation to solve multimodal learning tasks. Early works distinguished fusion approaches into feature-level early fusion and decision-level late fusion, depending on where fusion is performed in the model \cite{atrey2010multimodal}. \cite{Nefian2002DynamicBN} and \cite{Xu2018TexttoClipVR} adopted early fusion approach to integrating features from multiple modalities for speech recognition and video retrieval respectively. \cite{simonyan2014two} proposed to use two separate branches for spatial and temporal modalities and perform a simple late fusion for video action recognition. Alternatively, \cite{natarajan2012multimodal} fused the outputs by computing a weighted average. \cite{ye2012robust} proposed a robust late fusion using rank minimization. More recently, with the advancement of deep learning approaches, the idea of early fusion has been extended to the concept of mid fusion, where fusion happens at multiple levels \cite{ramachandram2017deep}. \cite{karpathy2014large} learned the fused representation by gradually fusing across multiple fusion layers. Similarly, \cite{vielzeuf2018centralnet} proposed a multilayer approach for fusion by introducing a central network linking all modality-specific networks. \cite{perez2019mfas} came up with an architecture search algorithm to find the optimal fusion architecture. \cite{hori2017attention,nagrani2021attention} incorporated attention mechanism for multimodal fusion. \cite{wang2020deep} proposed to exchange feature channels between modalities for multimodal fusion. \cite{pan2020x} introduced bilinear pooling to attention blocks, and demonstrated its superiority in capturing higher-level feature interactions by stacking multiple attention blocks for image captioning. More recently, attention has been moved to dynamic fusion, where the most suitable fusion strategy is selected from a set of candidate operations depending on input from different modalities \cite{han2022multimodal,xue2022dynamic}. Such dynamic approaches are more flexible to different multimodal tasks than static methods. Motivated by the success of capturing higher-level feature interactions and the dynamic fusion designs in multimodal fusion, our work aims to integrate the information within and across modalities at different levels by recursively applying nonlinear projections over intra- and inter-modality features, while generalizing well to multimodal tasks involving different modalities.

\textbf{Implicit Deep Learning} is a new family of deep neural networks and has grown rapidly in recent years. Traditional explicit deep models are often associated with a predefined architecture, and the backward pass is performed in reverse order through the explicit computation graphs. In contrast, implicit models compute their outputs by finding the root of some equations and analytically backpropagating through the root \cite{bai2020multiscale}. Previous works mainly focus on designing the hidden states of implicit models. \cite{pineda1987generalization} proposed an implicit backpropagation method for recurrent dynamics. \cite{amos2017optnet} proposed optimization layers to model implicit layers. Neural ODEs find the root of differentiable equations to model a recursive residual block \cite{chen2018neural}. Deep equilibrium models (DEQ) find a fixed point of the underlying system via black-box solvers, and are equivalent to going through an infinite depth feed-forward network \cite{bai2019deep,bai2020multiscale}. These implicit deep learning approaches have demonstrated competitive performance in multiple applications while vastly reducing memory consumption, \textit{e.g.}, generative models \cite{lu2021implicit,pokle2022deep}, optical flow \cite{teed2020raft,bai2022deep}, graph modeling \cite{li2021training}, \textit{etc}. \cite{bai2021stabilizing} also proposed a Jacobian regularization method to stabilize DEQs. Our work takes advantage of DEQs to adapt the number of recursion steps by finding the equilibrium state of intra- and inter-modality features jointly, and to speed up training and inference of our recursive fusion design. 

\begin{figure*}[tb]
\begin{center}
\centerline{\includegraphics[width=0.95\textwidth,page=4]{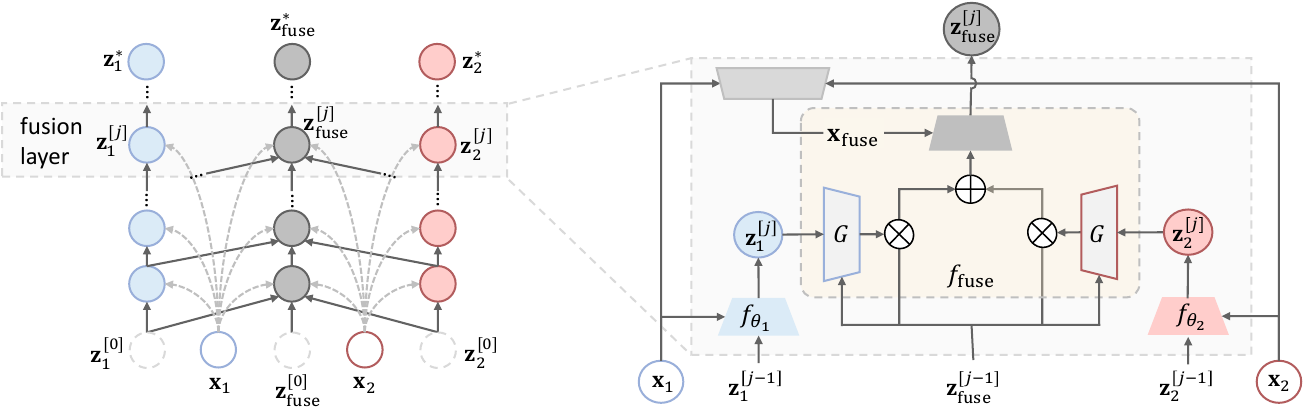}}
\caption{Our deep equilibrium fusion architecture. For simplicity, we illustrate the case where there are two modalities ($N=2$). The fusion layer is applied in a recursive manner until the equilibrium states are reached. Each layer $j$ computes its output based on the previous iteration. $\mathbf{z}^{[j]}$ denotes the output $\mathbf{z}$ at layer $j$. The modality-wise features $\mathbf{x}_1$ and $\mathbf{x}_2$ are injected at each layer, and are combined to obtain the residual fused feature $\mathbf{x}_{\mathrm{fuse}}$. $+$ represents summation and $\times$ denotes element-wise multiplication.}
\label{fig:deq-fusion}
\end{center}
\vspace{-8mm}
\end{figure*}

\section{Deep Equilibrium Fusion}
In this section, we first revisit the formulation of basic deep equilibrium models (DEQ) and then elaborate the formulation of our DEQ fusion for multimodal fusion.

\subsection{Revisiting Deep Equilibrium Model}\label{sec:deq}
Our DEQ fusion is particularly built on deep equilibrium models to recursively capture intra- and inter-modality interactions for multimodal fusion. The traditional deep neural networks have finite depth and perform the backward pass through every layer. Two interesting observations are that the hidden layers tend to converge to some fixed points, and employing the same weight in each layer of the network, so-called \emph{weight tying}, still achieves competitive results. That leads to the design principles of deep equilibrium models and the goal is to simulate an infinite depth weight-tied deep network, producing high-level and stable feature representations.

Formally, the standard DEQ~\cite{bai2019deep} is formulated as a weight-tied network, and such a network with parameter $\theta$ and a depth of $L$ computes a hidden state $\mathbf{z}$ as
\begin{equation}
    \mathbf{z}^{[j+1]}=f_\theta(\mathbf{z}^{[j]};\mathbf{x}),\quad j=0,\dots,L-1
\end{equation}
where the untransformed input $\mathbf{x}$ is injected at each layer, $\mathbf{z}^{[j]}$ is the hidden state at layer $j$ and $\mathbf{z}^{[0]}=\mathbf{0}$. As claimed in \cite{bai2019deep}, the core idea of DEQ is that when there are infinite layers ($L\rightarrow\infty$), the system tends to converge to an equilibrium state $\mathbf{z}^*$ such that
\begin{equation}\label{eq:deq-equilibrium-state}
    \mathbf{z}^*=f_\theta(\mathbf{z}^*;\mathbf{x}).
\end{equation}
In practice, naively computing the equilibrium state  requires excessive runtime. One convergence acceleration is to formulate \cref{eq:deq-equilibrium-state} into a root-finding problem: 
\begin{equation}
    g_\theta(\mathbf{z};\mathbf{x})=f_\theta(\mathbf{z};\mathbf{x})-\mathbf{z}.
\end{equation}
Some root solvers can then be applied to the residual $g_\theta$ to find the equilibrium state
\begin{equation}
    \mathbf{z}^*=\mathrm{RootSolver}(g_\theta;\mathbf{x}).
\end{equation}
Instead of backpropagating through each layer, we can compute gradients analytically as 
\begin{equation}\label{eq:deq-grad}
    \frac{\partial\ell}{\partial(\cdot)}=\frac{\partial\ell}{\partial\mathbf{z}^*}\left(\left.{-J_{g_\theta}^{-1}}\right|_{\mathbf{z}^*}\right)\frac{\partial f_\theta(\mathbf{z};\mathbf{x})}{\partial(\cdot)},
\end{equation}
where $\ell=\mathcal{L}(\mathbf{z}^*,\mathbf{y})$ is a loss between $\mathbf{z}^*$ and the target $\mathbf{y}$, $\left.{J_{g_\theta}^{-1}}\right|_{\mathbf{z}^*}$ is the inverse Jacobian of $g_\theta$ at $\mathbf{z}^*$. As it is expensive to compute the inverse Jacobian term, \cite{bai2019deep} proposed to alternatively solve a linear system by involving a vector-Jacobian product
\begin{equation}
    \mathbf{x}\left(\left.{J_{g_\theta}}\right|_{\mathbf{z}^*}\right)+\frac{\partial\ell}{\partial\mathbf{z}^*}=\mathbf{0}.
\end{equation}
With the formulation above, DEQ represents an infinite depth network with just one layer $f_\theta$, which converges to an equilibrium state, and can be backpropagated implicitly with a single computation.

\subsection{Deep Equilibrium Multimodal Fusion}
Next, we formulate our DEQ fusion method. Given a set of unimodal features $\mathbf{x}=\{\mathbf{x}_1,\mathbf{x}_2,\dots,\mathbf{x}_N\}$ from $N$ modalities, our goal is to find a unified feature that integrates the information from all modalities. To ensure the informativeness of our final integrated feature, we first execute another nonlinear projection $f_{\theta_i}(\cdot)$ to extract higher-level information within each modality:
\begin{equation}
    \mathbf{z}_i^{[j+1]}=f_{\theta_i}(\mathbf{z}_i^{[j]};\mathbf{x}_i),
\end{equation}
where $\mathbf{z}_i^{[j]}$ is the $j$-th output of the layer for modality $i$ and $\mathbf{z}_i^{[0]}$ is initialized to $\mathbf{0}$. $\mathbf{x}_i$ is the injected input feature for modality $i$. Our fusion design is flexible from the standpoint that $f_{\theta_i}(\cdot)$ can be altered arbitrarily to fit multiple modalities. In our case, $f_{\theta_i}(\cdot)$ is designed to be similar to a simple residual block \cite{he2016deep}. Following \cite{bai2020multiscale}, we adopt group normalization \cite{wu2018group} instead of batch normalization \cite{ioffe2015batch} for stability. Hence, $f_{\theta_i}(\cdot)$ is formulated as
\begin{equation}
\begin{gathered}
    \hat{\mathbf{z}}_i^{[j]}=\mathrm{ReLU}\left(\mathrm{GroupNorm}\left(\hat{\theta}_i\mathbf{z}_i^{[j]}+\hat{\mathbf{b}}_i\right)\right)\\
    \Tilde{\mathbf{z}}_i^{[j]}=\mathrm{GroupNorm}\left(\Tilde{\theta}_i\hat{\mathbf{z}}_i^{[j]}+\mathbf{x}_i+\Tilde{\mathbf{b}}_i\right)\\
    f_{\theta_i}(\mathbf{z}_i^{[j]};\mathbf{x}_i)=\mathrm{GroupNorm}\left(\mathrm{ReLU}\left(\Tilde{\mathbf{z}}_i^{[j]}\right)\right),
\end{gathered}
\end{equation}
where $\hat{\theta}_i$ and $\Tilde{\theta}_i$ are the weights, $\hat{\mathbf{b}}_i$ and $\Tilde{\mathbf{b}}_i$ are the bias. Given this set of modality-wise features $\{\mathbf{z}_i^{[j+1]}\}$ computed from $f_{\theta_i}(\cdot)$, where $i=1,2,\dots,N$, our target is to fuse them to obtain a unified feature integrating the information from all $N$ modalities. In addition, considering that the dimension of this unified feature is limited, it necessitates dynamically selecting the most representative information from each modality-wise feature to reduce redundancy. 

We propose a dynamic \emph{purify-then-combine} fusion strategy for this purpose. We account for feature correlation between the fused feature and the modality-wise features by applying a soft gating function $G(\cdot)$, to dynamically model feature correlation via computing a weight $\alpha_i$ for each modality:
\begin{equation}\label{eq:gating}
\begin{gathered}
    \alpha_i=G(\mathbf{z}_{\mathrm{fuse}}^{[j]},\mathbf{z}_i^{[j+1]})\\
    G(\mathbf{z}_{\mathrm{fuse}}^{[j]},\mathbf{z}_i^{[j+1]})={\theta_{\alpha}}\left(\mathbf{z}_{\mathrm{fuse}}^{[j]}+\mathbf{z}_i^{[j+1]}\right)+{\mathbf{b}_{\alpha}},
\end{gathered}
\end{equation}
where $\mathbf{z}_{\mathrm{fuse}}^{[j]}$ is the fused feature from the $j$-th layer and $\mathbf{z}_{\mathrm{fuse}}^{[0]}$ is initialized to $\mathbf{0}$. $\theta_{\alpha}$ and $\mathbf{b}_{\alpha}$ are the weight and bias. The gating function $G(\cdot)$ assigns the larger weights to parts of the fused feature that better encode the information from modality $i$. We purify the fused feature with the correlation weight for modality $i$:
\begin{equation}\label{eq:z-att}
    \mathbf{z}_i^{\prime}=\alpha_i\odot \mathbf{z}_{\mathrm{fuse}}^{[j]},
\end{equation}
where $\odot$ represents element-wise multiplication. $\mathbf{z}_i^{\prime}$ could be interpreted as the significant feature purified from the fused feature that represents the information of modality $i$ from previous layers. We then combine these purified features and adopt a simplified residual block to obtain the unified feature as
\begin{equation}\label{eq:z-fuse}
\begin{gathered}
    \hat{\mathbf{z}}_{\mathrm{fuse}}={\theta}_{\mathrm{fuse}}\cdot\sum_{i=1}^N \mathbf{z}_i^{\prime}+{\mathbf{b}}_{\mathrm{fuse}}\\
    \mathbf{z}_{\mathrm{fuse}}^{[j+1]}=\mathrm{GroupNorm}\left(\mathrm{ReLU}\left(\hat{\mathbf{z}}_{\mathrm{fuse}}+\mathbf{x}_{\mathrm{fuse}}\right)\right),
\end{gathered}
\end{equation}
where $\mathbf{x}_{\mathrm{fuse}}$ is the injected input fused feature computed from the set of modality-wise features $\{\mathbf{x}_i\}$ for $i=1,2,\dots,N$, $\theta_{\mathrm{fuse}}$ and $\mathbf{b}_{\mathrm{fuse}}$ are the weight and bias. In shallow layers (small $j$), $\mathbf{z}_{\mathrm{fuse}}^{[j]}$ encodes low-level modality interactions. As we continuously summarize the purified feature $\mathbf{z}_i^{\prime}$, \textit{i.e.}, $j$ gets larger and larger, $\mathbf{z}_{\mathrm{fuse}}^{[j]}$ tends to capture higher-level modality interactions while recursively integrating low-level information from previous iterations. By doing so, the final $\mathbf{z}_{\mathrm{fuse}}^{[\infty]}$ integrates the cross-modality interactions and correlations ranging from low level to high level. Moreover, our approach is flexible on the ways to compute the injected fused feature $\mathbf{x}_{\mathrm{fuse}}$. In our case, we compute it with a simple weighted sum:
\begin{equation}
    \mathbf{x}_{\mathrm{fuse}}=\sum_{i=1}^N w_i\mathbf{x}_i,
\end{equation}
where $w_i$ is a learnable weight associated with modality $i$ representing modality importance.

We denote the above-proposed fusion module in \cref{eq:gating,eq:z-att,eq:z-fuse} as a nonlinear function $f_{\mathrm{fuse}}(\cdot)$ such that
\begin{equation}
    \mathbf{z}_{\mathrm{fuse}}^{[j+1]}=f_{\mathrm{fuse}}(\mathbf{z}_{\mathrm{fuse}}^{[j]};\mathbf{x}),
\end{equation}
where $\mathbf{x}=\{\mathbf{x}_i\}$ for $i=1,2,\dots,N$ is the set of the injected modality-wise features. Ideally, a superior unified feature should capture the information from all modalities at every level and thus we progressively model modality interactions from low-level to high-level feature space. Technically, we present to recursively interchange intra- and inter-modality information until the equilibrium state is reached, to obtain such an informative unified representation in a \emph{stable} feature space for multimodal learning. To achieve this goal, we leverage the idea of DEQs into our multimodal fusion framework. Considering $f_{\theta_i}(\cdot)$ for $i=1,2,\dots,N$ and $f_{\mathrm{fuse}}(\cdot)$ as DEQ layers, we aim to find equilibrium states such that 
\begin{equation}\label{eq:fixed-point}
    \mathbf{z}_i^{*}=f_{\theta_i}\left(\mathbf{z}_i^{*};\mathbf{x}_i\right),\quad\mathbf{z}_{\mathrm{fuse}}^{*}=f_{\mathrm{fuse}}\left(\mathbf{z}_{\mathrm{fuse}}^{*};\mathbf{x}\right),
\end{equation}
where $\mathbf{z}_{\mathrm{fuse}}^{*}$ and $\mathbf{z}_i^{*}$, $i=1,2,\dots,N$, are the fused feature and all unimodal features in equilibrium states respectively. Note that we also keep track of computation for each unique modality-wise feature, so that the information from different modalities can be aligned and captured at a stable level together with the fused feature. We conduct ablation studies to demonstrate the superiority of our \emph{purify-then-combine} fusion strategy compared to other fusion variants involving DEQs. Please refer to \cref{sec:ablation} for more details.

The fixed points in \cref{eq:fixed-point} can be reformulated into residual functions for the root-finding problem:
\begin{equation}\label{eq:residual-g-theta}
    g_{\theta_i}(\mathbf{z}_i;\mathbf{x}_i)=f_{\theta_i}(\mathbf{z}_i;\mathbf{x}_i)-\mathbf{z}_i,
\end{equation}
\begin{equation}\label{eq:residual-g-fuse}
    g_{\mathrm{fuse}}(\mathbf{z}_{\mathrm{fuse}};\mathbf{x})=f_{\mathrm{fuse}}(\mathbf{z}_{\mathrm{fuse}};\mathbf{x})-\mathbf{z}_{\mathrm{fuse}}
\end{equation}
Finally, we can solve for features in equilibrium states via a black-box solver by minimizing the residuals $g_{\theta_i}$ for $i=1,2,\dots,N$ and $g_{\mathrm{fuse}}$:
\begin{equation}
    \mathbf{z}^{*},\mathbf{z}_{\mathrm{fuse}}^*=\mathrm{RootSolver}(g_{\theta};g_{\mathrm{fuse}};\mathbf{x}),
\end{equation}
where $\mathbf{z}^{*}=\{\mathbf{z}_i^{*}\}$ and $g_{\theta}=\{g_{\theta_i}\}$ for $i=1,2,\dots,N$. \cref{fig:deq-fusion} illustrates an overview of our deep equilibrium fusion architecture. 

\subsection{Backpropagation}
A benefit of using DEQs compared to stacking conventional networks is that the gradients can be computed analytically without tracing through the forward pass layer-by-layer.
\begin{theorem}\label{thm:loss-x-grad}\textbf{(Gradient of Deep Equilibrium Multimodal Fusion)}
Let $\mathbf{z}_i^*,\mathbf{z}_{\mathrm{fuse}}^{*}\in \mathbb{R}^{d}$ for $i=1,2,\dots,N$ be the equilibrium states of the modality-wise features and fused feature, and $\mathbf{y}\in \mathbb{R}^{q}$ be the ground-truth. Suppose we have a function $h:\mathbb{R}^d\rightarrow\mathbb{R}^q$ which is the head for some downstream tasks (\textit{e.g.}, classification), we can compute a loss function $\ell=\mathcal{L}(h(\mathbf{z}_{\mathrm{fuse}}^{*}),\mathbf{y})$ between the prediction and the target. We can backpropagate implicitly through the unimodal features by computing the gradients with respect to $\mathbf{x}_i$ using implicit function theorem:
\begin{equation}
\begin{aligned}
    \frac{\partial\ell}{\partial\mathbf{x}_i}=\frac{\partial\ell}{\partial\mathbf{z}_{\mathrm{fuse}}^{*}}\cdot\left(\left.{-J_{g_{\mathrm{fuse}}}^{-1}}\right|_{\mathbf{z}_{\mathrm{fuse}}^{*}}\right)\cdot\frac{\partial f_{\mathrm{fuse}}\left(\mathbf{z}_{\mathrm{fuse}}^{*};\mathbf{x}\right)}{\partial\mathbf{z}_i^*}\cdot\left({-J_{g_{\theta_i}}^{-1}}|_{\mathbf{z}_{i}^{*}}\right)\cdot\frac{\partial f_{\theta_i}\left(\mathbf{z}_{i}^{*};\mathbf{x}_{i}\right)}{\partial\mathbf{x}_i},
\end{aligned}
\end{equation}
where $\left.{J_{g}^{-1}}\right|_{\mathbf{z}}$ is the inverse Jacobian of $g$ evaluated at $\mathbf{z}$.   
\end{theorem}
The proof for \cref{thm:loss-x-grad} is provided in \cref{appendix:proof-backward}. The gradients with respect to parameters of DEQ layers can be computed following \cref{eq:deq-grad}.

\begin{figure}[t]
\begin{center}
\includegraphics[width=\textwidth]{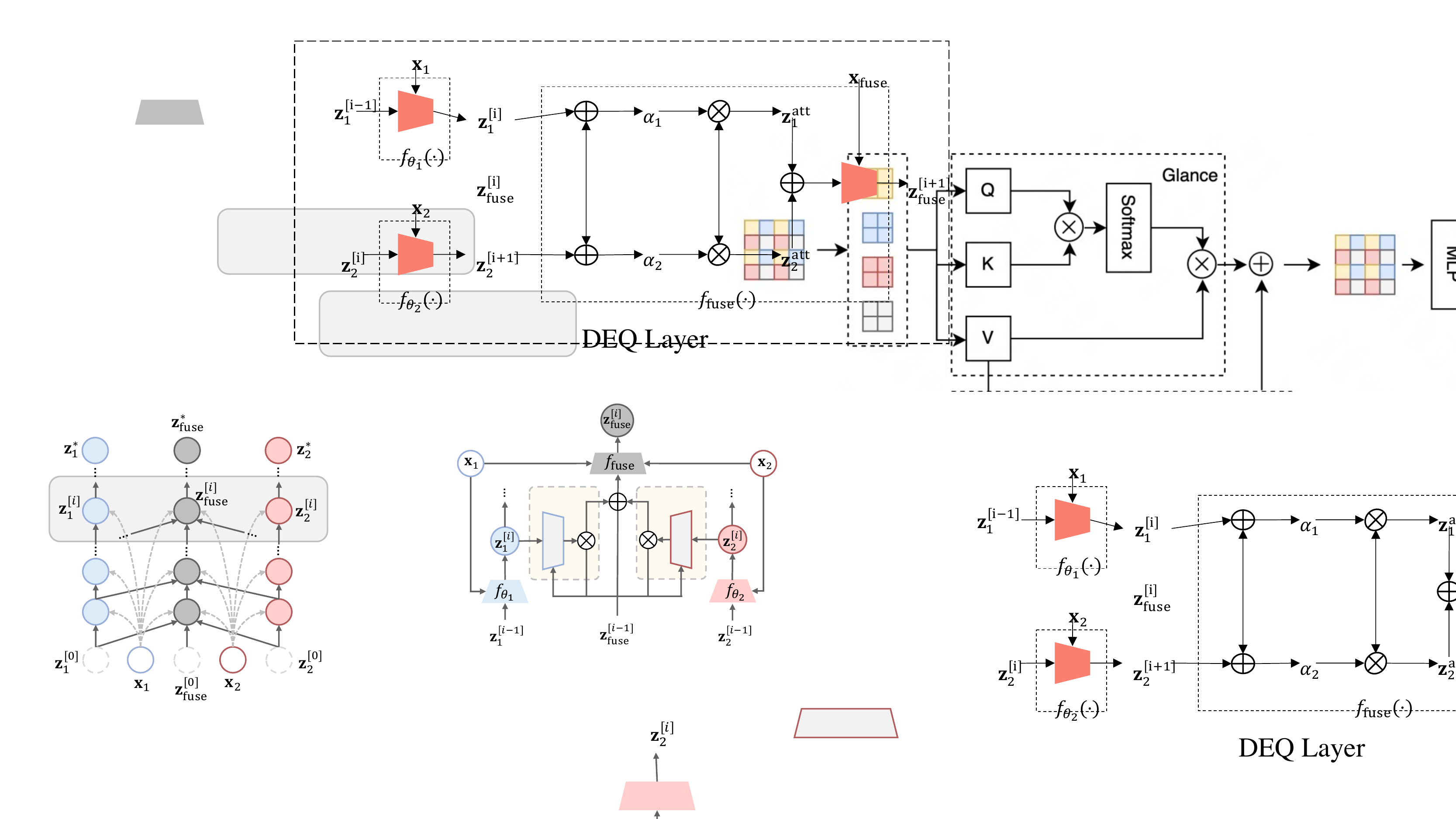}
\caption{Data samples from the FIVE benchmarks: (a) multi-omics BRCA; (b) image-text MM-IMDB; (c) audio-text CMU-MOSI; (d) image-point SUN RGB-D; and (e) image-question VQA-v2.}
\label{fig:data-sample}
\end{center}
\end{figure}

\section{Experiments}
We empirically verify the merit of our DEQ fusion on five multimodal tasks: 1) breast invasive carcinoma PAM50 subtype classification BRCA\footnote{BRCA can be acquired from \href{https://www.cancer.gov/about-nci/organization/ccg/research/structural-genomics/tcga}{The Cancer Genome Atlas program}.}, associated with mRNA expression, DNA methylation, and miRNA expression data; 2) movie genre classification on MM-IMDB \cite{arevalo2017gated}, which categorizes movies based on posters and text descriptions; 3) sentiment analysis on CMU-MOSI \cite{zadeh2016mosi}, which manually labels sentiment of video clips ranging from -3 to 3, where -3 indicates highly negative and 3 indicates highly positive; 4) 3D object detection on SUN RGB-D \cite{song2015sun}, one of the most challenging large-scale benchmarks for regressing 3D object bounding bbox offsets and predicting its category; and 5) visual question answering on VQA-v2 \cite{goyal2017making}, the most commonly used large-scale VQA benchmark dataset containing human-annotated question-answer relating to images.
\cref{fig:data-sample} illustrates some data examples. In order to demonstrate the generalizability and plug-and-play nature of our approach, we only replace the fusion module of the existing methods and remain all the other components the same for comparison. The detailed experimental setup is demonstrated in \cref{sec:setup}.

\subsection{Discussion}

\begin{table*}[t]
\caption{Performance comparisons of multimodal fusion methods on BRCA benchmark. The results of baseline methods are obtained from \cite{han2022multimodal}. mR, D, and miR denote mRNA expression, DNA methylation, and miRNA expression data respectively. $\uparrow$ indicates the higher the metric the better the performance and vice versa for $\downarrow$. The best results are in bold.}\label{table:brca}
\small
\centering
\begin{tabular}{lcccc}
\toprule
         & Modality & Acc(\%)$\uparrow$ & WeightedF1(\%)$\uparrow$ & MacroF1(\%)$\uparrow$ \\
\midrule
GRridge \cite{van2016better} & mR+D+miR & 74.5$\pm$1.6 & 72.6$\pm$2.5 & 65.6$\pm$2.5 \\
GMU \cite{arevalo2017gated} & mR+D+miR & 80.0$\pm$3.9 & 79.8$\pm$5.8 & 74.6$\pm$5.8 \\
CF \cite{hong2020more} & mR+D+miR & 81.5$\pm$0.8 & 81.5$\pm$0.9 & 77.1$\pm$0.9 \\
MOGONET \cite{wang2021mogonet} & mR+D+miR & 82.9$\pm$1.8 & 82.5$\pm$1.7 & 77.4$\pm$1.7 \\
TMC \cite{han2021trusted} & mR+D+miR & 84.2$\pm$0.5 & 84.4$\pm$0.9 & 80.6$\pm$0.9 \\
\midrule
MM-Dynamics \cite{han2022multimodal} & mR+D+miR & 87.7$\pm$0.3 & 88.0$\pm$0.5 & 84.5$\pm$0.5 \\
MM-Dynamics + \textbf{DEQ Fusion} & D+miR & 78.9$\pm$1.6 & 79.2$\pm$2.3 & 75.8$\pm$3.0 \\
MM-Dynamics + \textbf{DEQ Fusion} & mR+miR & 87.6$\pm$0.7 & 88.1$\pm$0.7 & 85.1$\pm$1.7 \\
MM-Dynamics + \textbf{DEQ Fusion} & mR+D & 88.7$\pm$0.7 & 89.3$\pm$0.7 & 86.9$\pm$0.9 \\
MM-Dynamics + \textbf{DEQ Fusion} & mR+D+miR & \textbf{89.1$\pm$0.7} & \textbf{89.7$\pm$0.7} & \textbf{87.6$\pm$1.0} \\
\bottomrule
\end{tabular}
\end{table*}

\textbf{BRCA.}\quad We compare our DEQ fusion approach with several baseline fusion methods, including the best competitor MM-Dynamics \cite{han2022multimodal}, in \cref{table:brca}. 
It is noticeable that the complementarity of some modalities is significant, as approximately -10\% performance drop is observed without mRNA data.
This also somewhat manifests the advantage of dynamic modeling to take multiple modality signals into account. Similar to our dynamic design with a soft gating function, MM-Dynamics models feature and modality informativeness dynamically for trustworthy multimodal fusion.
Our DEQ fusion additionally considers intra- and inter-modality features at every level, outperforming MM-Dynamics in all evaluation metrics. Notably, our method with two modalities of mRNA and DNA methylation already attains better performance in all evaluation metrics compared to MM-Dynamics which leverages all three modalities. All above results demonstrate the effectiveness of capturing modality interactions ranging from low level to high level in our deep equilibrium fusion design.

\begin{wraptable}{o}{0.62\textwidth}
\caption{Performance comparisons of multimodal fusion methods on MM-IMDB benchmark. 
The result of DynMM is obtained from \cite{xue2022dynamic}. I and T denote image and text respectively. 
}
\label{table:mm-imdb}
\centering
\small
\begin{tabular}{lccc}
\toprule
         & Modality & MicroF1(\%)$\uparrow$ & MacroF1(\%)$\uparrow$ \\
\midrule
Unimodal Image & I & 40.31 & 25.76 \\
Unimodal Text & T & 59.37 & 47.59 \\
Early Fusion & I+T & 56.00 & 49.36 \\
LRMF \cite{liu2018efficient} & I+T & 58.95 & 50.73 \\
MFM \cite{tsai2018learning} & I+T & 56.44 & 48.53 \\
MI-Matrix \cite{jayakumar2020multiplicative} & I+T & 55.87 & 46.77 \\
RMFE \cite{gat2020removing} & I+T & 58.67 & 49.82 \\
CCA \cite{sun2020learning} & I+T & 60.31 & 50.45 \\
RefNet \cite{sankaran2021multimodal} & I+T & 59.45 & 51.51 \\
DynMM \cite{xue2022dynamic} & I+T & 60.35 & 51.60 \\
\midrule
Late Fusion & I+T & 59.02 & 50.27 \\
\textbf{DEQ Fusion} & I+T & \textbf{61.52} & \textbf{53.38} \\
\bottomrule
\end{tabular}
\vspace{-4mm}
\end{wraptable}

\textbf{MM-IMDB.}\quad We compare our DEQ fusion strategy with various baseline fusion methods in \cref{table:mm-imdb}. It is clear that text modality is more representative than image modality for this classification task, as unimodal text models exhibit significantly better performance than unimodal image models. As such, existing approaches
which do not involve dynamic modeling of modality information, attain either similar performance or minor improvement compared to the unimodal text baseline. A dynamic fusion strategy is seemingly crucial to further leverage the information from the relatively weak image signal for better performance. DynMM~\cite{xue2022dynamic} capitalizes on hard gating to select the most appropriate fusion strategy from a set of predefined operations
to achieve better results. We experiment with a late fusion strategy by simply replacing the original concatenation fusion with our DEQ fusion module.
With this simple modification, we obtain the state-of-the-art results of 61.52\% and 53.38\% for micro and macro F1 scores respectively on MM-IMDB benchmark, which is a significant improvement of 2.50\% and 3.11\% against the late fusion baseline, also 1.17\% and 1.78\% improvement compared to DynMM.

\begin{table*}[b]
\vspace{-4mm}
\caption{Performance comparisons of multimodal fusion methods on CMU-MOSI benchmark. 
The results of baseline methods are obtained from \cite{yang2020cm}. T, A, and V denote text, audio, and video, respectively. Acc-$N$ denotes $N$-class accuracy.} 
\label{table:cmu-mosi}
\centering
\small
\begin{tabular}{lcccccc}
\toprule
         & Modality & Acc-7(\%)$\uparrow$ & Acc-2(\%)$\uparrow$ & F1(\%)$\uparrow$ & MAE$\downarrow$ & Corr$\uparrow$ \\
\midrule
Early Fusion LSTM & T+A+V & 33.7 & 75.3 & 75.2 & 1.023 & 0.608 \\
LRMF \cite{liu2018efficient} & T+A+V & 32.8 & 76.4 & 75.7 & 0.912 & 0.668 \\
MFN \cite{zadeh2018memory} & T+A+V & 34.1 & 77.4 & 77.3 & 0.965 & 0.632 \\
MARN \cite{zadeh2018multi} & T+A+V & 34.7 & 77.1 & 77.0 & 0.968 & 0.625 \\
RMFN \cite{liang2018multimodal} & T+A+V & 38.3 & 78.4 & 78.0 & 0.922 & 0.681 \\
MFM \cite{tsai2018learning} & T+A+V & 36.2 & 78.1 & 78.1 & 0.951 & 0.662 \\
MCTN \cite{pham2019found} & T+A+V & 35.6 & 79.3 & 79.1 & 0.909 & 0.676 \\
MulT \cite{tsai2019multimodal} & T+A+V & 40.0 & 83.0 & 82.8 & 0.871 & 0.698 \\
\midrule
BERT \cite{devlin2018bert} & T & 41.5 & 83.2 & 82.3 & 0.784 & 0.774 \\
CM-BERT \cite{yang2020cm} & T+A & 44.9 & 84.5 & 84.5 & \textbf{0.729} & 0.791 \\
CM-BERT + \textbf{DEQ Fusion} & T+A & \textbf{46.1} & \textbf{85.4} & \textbf{85.4} & 0.737 & \textbf{0.797} \\
\bottomrule
\end{tabular}
\vspace{-2mm}
\end{table*}

\textbf{CMU-MOSI.}\quad We compare our fusion approach with several baseline fusion methods, including the state-of-the-art CM-BERT \cite{yang2020cm}, in \cref{table:cmu-mosi}. It is worth noting that BERT-based methods exhibit better performance than other baseline approaches. For instance, vanilla BERT \cite{devlin2018bert}, leveraging only text modality, already surpasses other non-BERT methods which involve the utilization of all three modalities. We speculate that text modality provides more significant information for sentiment analysis task than the other two modalities. CM-BERT exploits audio modality in addition to BERT for further performance boost. Our DEQ fusion benefits from the dynamic and stable modality information modeling, and interaction exchange at every level with our recursive fusion design, outperforming CM-BERT by 1.2\%, 0.9\%, and 0.9\% in Acc7, Acc2, and F1 score, respectively. 

\begin{table*}[t]
\caption{Performance comparisons of multimodal fusion methods on SUN RGB-D benchmark. P denotes point cloud and H denotes height. \emph{repro.} denotes our reproduced results.}
\label{table:sun-rgbd}
\small
\centering
\begin{tabular}{lcccc}
\toprule
Method + \emph{Fusion Method} & Modality & mAP@0.25 & mAP@0.5 & Gain on mAP@0.25\\
\midrule
GroupFree~\cite{liu2021group} & P & 63.0 & 45.2 & - \\
GroupFree~\cite{liu2021group} + Simple Appending & P+RGB & 62.1 & 42.7 & -0.5 \\
\midrule
VoteNet~\cite{qi2019deep} & P & 57.7 & - & - \\
VoteNet~\cite{qi2019deep} + Simple Appending & P+RGB & 56.3 & - & -1.4 \\
VoteNet~\cite{qi2019deep} + TupleInfoNCE~\cite{liu2021contrastive} & P+RGB+H & 58.0 & - & +0.3 \\
\midrule
ImVoteNet~\cite{qi2020imvotenet} & P+RGB & \textbf{63.4} & - & - \\
ImVoteNet~\cite{qi2020imvotenet} \emph{repro.} & P+RGB & 61.9 & 45.6 & - \\
ImVoteNet~\cite{qi2020imvotenet} \emph{repro.} + \textbf{DEQ Fusion} & P+RGB & 62.7 & \textbf{46.4} & \textbf{+0.8} \\
\bottomrule
\end{tabular}
\end{table*}

\textbf{SUN RGB-D.}\quad We report mean Average Precision (mAP) with 3D IoU thresholds of 0.25 and 0.5 measured on multiple 3D object detection methods in \cref{table:sun-rgbd}. Interestingly, adding the additional RGB modality without advanced fusion mechanism harms the performance, \emph{e.g.}, including RGB modality into GroupFree~\cite{liu2021group} and VoteNet~\cite{qi2019deep} with simple appending fusion leads to -0.5\% and -1.4\% performance drop. This is a strong indication of the difficulty in fusing useful RGB information into the extensive point cloud information. TupleInfoNCE~\cite{liu2021contrastive} designs a contrastive loss for multimodal representation learning, and contributes to a performance gain of +0.3\% on mAP@0.25 from VoteNet baseline with additional RGB and height modalities. In addition to VoteNet, ImVoteNet~\cite{qi2020imvotenet} additionally proposes image votes to boost 3D object detection performance. By plugging our DEQ fusion into ImVoteNet, we obtain +0.8\% gain on mAP@0.25 compared to ImVoteNet baseline. Note that the performance of our reproduced ImVoteNet (ImVoteNet \emph{repro.}) is slightly lower than the one reported in the original paper, and our experiments are based on our reproduced implementation.

\begin{table*}[t]
\caption{Performance comparisons of multimodal fusion methods on VQA-v2 benchmark. All metrics are accuracy in \%.}
\label{table:vqa}
\small
\centering
\begin{tabular}{lccccc}
\toprule
Basic Settings & Fusion Method & Yes/no & Number & Other & Overall \\
\midrule
Skip-thoughts + BottomUp & Mutan~\cite{ben2017mutan} & 82.40 & 42.63 & 54.85 & 63.73 \\
Skip-thoughts + BottomUp & \textbf{DEQ Fusion} & \textbf{82.91} & \textbf{45.40} & \textbf{55.70} & \textbf{64.57} \\
\midrule
GloVe + BottomUp + Self-Att + Guided-Att & MCAN~\cite{yu2019deep} & 84.67 & 48.44 & 58.52 & 67.02 \\
GloVe + BottomUp + Self-Att + Guided-Att & \textbf{DEQ Fusion} & \textbf{85.17} & \textbf{49.07} & \textbf{58.69} & \textbf{67.38} \\
\bottomrule
\end{tabular}
\vspace{-2mm}
\end{table*}

\textbf{VQA-v2.}\quad Our experimental results on VQA-v2 based on Mutan~\cite{ben2017mutan} and MCAN~\cite{yu2019deep} are shown in \cref{table:vqa}. Mutan~\cite{ben2017mutan} initializes GRU with pretrained Skip-thoughts models~\cite{kiros2015skip} to process questions, whereas MCAN~\cite{yu2019deep} leverages pretrained GloVe word embeddings~\cite{pennington2014glove}. Both methods use bottom-up attention visual features. In addition, MCAN introduces self-attention and guided-attention units to model intra- and inter-modality interactions. Following their basic settings, we replace the fusion method with our DEQ fusion for comparison. We achieve consistent improvements over all evaluation metrics on both baselines, suggesting the superiority of our method.

\begin{table}[t]
\begin{minipage}[t]{0.55\textwidth}
\centering
\caption{Ablation experiments on BRCA. $f_{\theta}$ represents the modality-wise nonlinear projections $f_{\theta_i}(\cdot)$ for $i=1,2,\dots,N$; $f_{\mathrm{fuse}}$ denotes the fusing function $f_{\mathrm{fuse}}(\cdot)$; \emph{DEQ} indicates enabling recursive DEQ computation to find the equilibrium state for the functions.} 
\label{table:ablation}
\vskip 0.1in
\centering
\small
\begin{tabular}{cccccc}
\toprule
& & & & \multicolumn{2}{c}{F1(\%)$\uparrow$} \\
$f_{\theta}$ & $f_{\mathrm{fuse}}$ & DEQ & Acc(\%)$\uparrow$ & Weighted & Macro \\
\midrule
 & & & 87.6$\pm$0.4 & 87.9$\pm$0.4 & 84.3$\pm$0.8 \\
 $\checkmark$ & $\checkmark$ & & 86.2$\pm$0.6 & 86.5$\pm$0.6 & 82.9$\pm$0.9 \\
$\checkmark$ & & $\checkmark$ & 88.8$\pm$0.4 & 89.4$\pm$0.4 & 87.2$\pm$0.8 \\
 & $\checkmark$ & $\checkmark$ & 88.3$\pm$0.5 & 88.8$\pm$0.5 & 86.0$\pm$1.0 \\
$\checkmark$ & $\checkmark$ & $\checkmark$ & \textbf{89.1$\pm$0.7} & \textbf{89.7$\pm$0.7} & \textbf{87.6$\pm$1.0} \\
\bottomrule
\end{tabular}
\vspace{-4mm}
\end{minipage}\hfill
\begin{minipage}[t]{0.42\linewidth}
\vskip 0.1in
\begin{center}
\centerline{\includegraphics[width=\textwidth]{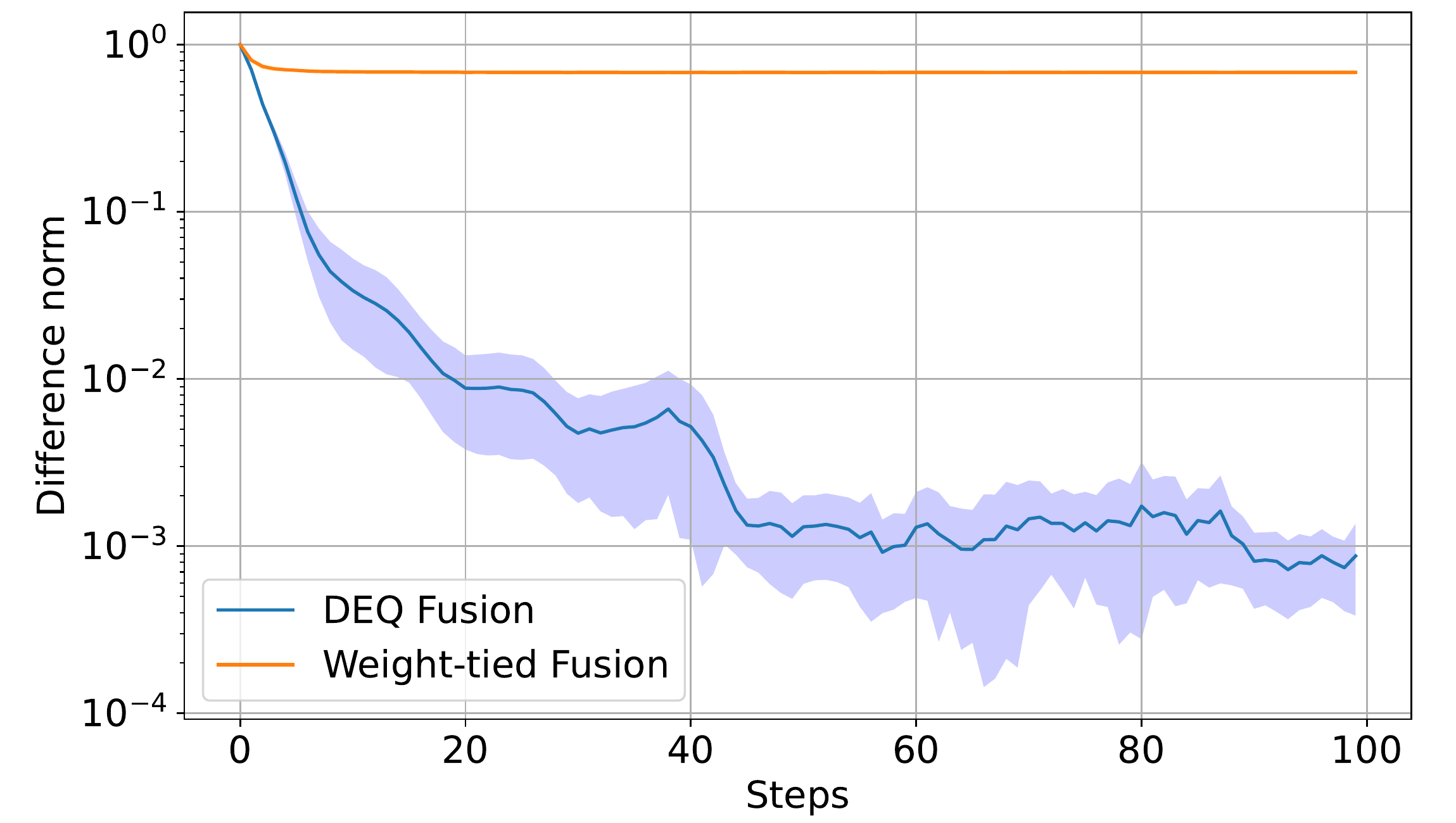}}
\vskip 0.1in
\captionof{figure}{Plot of DEQ Fusion's convergence to equilibrium over 100 solver steps. The shaded region indicates the 95\% confidence interval computed over 10 runs.}
\label{fig:converge}
\end{center}
\vskip -0.3in
\end{minipage}
\vspace{-10mm}
\end{table}


\subsection{Ablation Studies}\label{sec:ablation}
We conduct extensive ablation experiments to study the effectiveness of our proposed deep equilibrium fusion method from different perspectives. 
\cref{table:ablation} details the results. 
All ablation studies are evaluated on BRCA benchmark using all three modalities, following the same experimental setup stated in \cref{sec:setup}. Additional ablation studies on other benchmarks are in \cref{sec:add-ablation}.

\textbf{Effectiveness of seeking equilibrium.}\quad
We first examine the effectiveness of computing the equilibrium state to extract and integrate stable modality information at every level.
We first discard all components, \textit{i.e.}, directly fusing with a weighted sum approach: $\mathbf{x}_{\mathrm{fuse}}=\sum_{i=1}^N w_i\mathbf{x}_i$, where $w_i$ is a learnable weight associated with modality $i$. As shown in \cref{table:ablation}, this baseline fusion method obtains similar performance to \cite{han2022multimodal}. Next, we disable the recursive computation in our DEQ fusion module, \textit{i.e.}, all $f_{\theta_i}(\cdot)$ and $f_{\mathrm{fuse}}(\cdot)$ are only applied once without finding the equilibrium states. Since all inputs $\mathbf{z}$ are initialized to zero, this approach is equivalent to the weighted sum approach but with an additional nonlinear projection $f_{\theta_i}(\cdot)$ applied to all modality-wise features. Interestingly, introducing additional parameters without DEQ even harms performance compared to the weighted sum baseline. Results from both ablation studies demonstrate the importance of seeking the equilibrium states for multimodal fusion.


\textbf{Different fusion variants involving DEQ.}\quad
We compare our DEQ fusion strategy against several variants involving DEQ in \cref{table:ablation}.
First, we 
disable the \emph{purified-then-combine} fusion strategy, \emph{i.e.}, ablating our fusing projection $f_{\mathrm{fuse}}(\cdot)$ by simply summating all modality-wise features: $\mathbf{z}_{\mathrm{fuse}}^*=\sum_{i}^N \mathbf{z}_i^*$. 
Our full DEQ fusion notably improves all evaluation metrics compared to the runs without the proposed \emph{purified-then-combine} fusion strategy.
Next, we 
ablate all modality projections $f_{\theta_i}(\cdot)$ as identity functions by setting $\mathbf{z}_i^*=\mathbf{x}_i$. 
Specifically, given a set of features from $N$ modalities $\{\mathbf{x}_i\},i=1,2,\dots,N$, we set $\mathbf{z}_i^*=\mathbf{x}_i$.
and proceed fusion with $f_{\mathrm{fuse}}(\cdot)$.
We notice a decline in all evaluation metrics without modality-wise nonlinear projections.
These studies demonstrate that our proposed fusion variant produces the most encouraging results across all evaluation metrics.

\begin{table}[t]
\centering
\caption{Ablation experiments of soft gating function on BRCA. $G(\cdot)$ denotes the soft gating function.} 
\label{table:ablation-gating}
\vskip 0.1in
\centering
\small
\begin{tabular}{ccccccc}
\toprule
$f_{\theta}$ & $f_{\mathrm{fuse}}$ & DEQ & $G(\cdot)$ & Acc(\%)$\uparrow$ & WeightedF1(\%)$\uparrow$ & MacroF1(\%)$\uparrow$ \\
\midrule
$\checkmark$ & $\checkmark$ & $\checkmark$ &  & 88.4$\pm$0.8 & 89.0$\pm$0.8 & 86.1$\pm$1.1\\
$\checkmark$ & $\checkmark$ & $\checkmark$ & $\checkmark$ & \textbf{89.1$\pm$0.7} & \textbf{89.7$\pm$0.7} & \textbf{87.6$\pm$1.0} \\
\bottomrule
\end{tabular}
\vspace{-4mm}
\end{table}

\textbf{Impact of soft gating function.}\quad Motivated by the success of dynamically perceiving information from modalities, we develop a soft gating function to capture the important information within each modality. We further validate the effectiveness of the proposed soft gating function $G(\cdot)$. Specifically, we set $\mathbf{z}_i^{\prime}=\mathbf{z}_i^{[j+1]}$ for \cref{eq:z-att} to disable the soft gating function. As shown in \cref{table:ablation-gating}, DEQ fusion without soft gating function leads to about -1\% performance drop among all evaluation metrics. Note that since $G(\cdot)$ is a part of $f_{\mathrm{fuse}}$, disabling $f_{\mathrm{fuse}}$ automatically removes $G(\cdot)$. The soft gating function combined with all other components leads to the most superior result.

\textbf{Convergence of DEQ Fusion.}\quad
We examine the convergence of our DEQ fusion, which is an important assumption since fusion may collapse if it fails to find the equilibrium.
We train a model with our DEQ fusion from scratch, and track the relative difference norm evaluated as ${\|\mathbf{z}_{\mathrm{fuse}}^{[i+1]}-\mathbf{z}_{\mathrm{fuse}}^{[i]}\|}/{\|\mathbf{z}_{\mathrm{fuse}}^{[i]}\|}$ over 100 solver steps during inference. We compare it with a weight-tied fusion approach which simply iterates our fusion layer and performs backward pass layer-by-layer. \cref{fig:converge} depicts the empirical results. It is notable that the difference norm of our DEQ fusion quickly drops below 0.01 on average within 20 solver steps, whereas the weight-tied fusion oscillates around a relatively high value. Benefiting from fixed point solvers and analytical backward pass, our DEQ fusion has much quicker and stabler convergence to the fixed point than the weight-tied approach. 


\section{Conclusion}
We have presented an adaptive deep equilibrium (DEQ) approach for multimodal fusion. Our approach recursively captures intra- and inter-modality feature interactions until an equilibrium state is reached, encoding cross-modal interactions ranging from low level to high level for effective downstream multimodal learning. This deep equilibrium approach can be readily pluggable into existing multimodal learning frameworks to obtain further performance gain. 
More remarkably, our DEQ fusion constantly achieves new state-of-the-art performances on multiple multimodal benchmarks, showing its high generalizability and extendability. A common drawback of DEQ in applications is its additional training costs for solving root-finding and uncertain computation costs during inference. Although accelerating DEQ training and inference is not a focus of this work, improving the convergence of DEQs is an important direction, which we leave as future works.




{
\small
\bibliographystyle{plain}
\bibliography{deq_fusion}

\begin{thebibliography}{10}

\bibitem{amos2017optnet}
Brandon Amos and J~Zico Kolter.
\newblock Optnet: Differentiable optimization as a layer in neural networks.
\newblock In {\em International Conference on Machine Learning}, pages
  136--145. PMLR, 2017.

\bibitem{anderson1965iterative}
Donald~G Anderson.
\newblock Iterative procedures for nonlinear integral equations.
\newblock {\em Journal of the ACM (JACM)}, 12(4):547--560, 1965.

\bibitem{arevalo2017gated}
John Arevalo, Thamar Solorio, Manuel Montes-y G{\'o}mez, and Fabio~A
  Gonz{\'a}lez.
\newblock Gated multimodal units for information fusion.
\newblock {\em arXiv preprint arXiv:1702.01992}, 2017.

\bibitem{atrey2010multimodal}
Pradeep~K Atrey, M~Anwar Hossain, Abdulmotaleb El~Saddik, and Mohan~S
  Kankanhalli.
\newblock Multimodal fusion for multimedia analysis: a survey.
\newblock {\em Multimedia Systems}, 16(6):345--379, 2010.

\bibitem{bai2022deep}
Shaojie Bai, Zhengyang Geng, Yash Savani, and J~Zico Kolter.
\newblock Deep equilibrium optical flow estimation.
\newblock In {\em Proceedings of the IEEE/CVF Conference on Computer Vision and
  Pattern Recognition}, pages 620--630, 2022.

\bibitem{bai2019deep}
Shaojie Bai, J~Zico Kolter, and Vladlen Koltun.
\newblock Deep equilibrium models.
\newblock {\em Advances in Neural Information Processing Systems}, 32, 2019.

\bibitem{bai2020multiscale}
Shaojie Bai, Vladlen Koltun, and J~Zico Kolter.
\newblock Multiscale deep equilibrium models.
\newblock {\em Advances in Neural Information Processing Systems},
  33:5238--5250, 2020.

\bibitem{bai2021stabilizing}
Shaojie Bai, Vladlen Koltun, and J~Zico Kolter.
\newblock Stabilizing equilibrium models by jacobian regularization.
\newblock {\em arXiv preprint arXiv:2106.14342}, 2021.

\bibitem{banos2015design}
Oresti Banos, Claudia Villalonga, Rafael Garcia, Alejandro Saez, Miguel Damas,
  Juan~A Holgado-Terriza, Sungyong Lee, Hector Pomares, and Ignacio Rojas.
\newblock Design, implementation and validation of a novel open framework for
  agile development of mobile health applications.
\newblock {\em Biomedical Engineering Online}, 14(2):1--20, 2015.

\bibitem{ben2017mutan}
Hedi Ben-Younes, R{\'e}mi Cadene, Matthieu Cord, and Nicolas Thome.
\newblock Mutan: Multimodal tucker fusion for visual question answering.
\newblock In {\em Proceedings of the IEEE International Conference on Computer
  Vision}, pages 2612--2620, 2017.

\bibitem{chen2018neural}
Ricky~TQ Chen, Yulia Rubanova, Jesse Bettencourt, and David~K Duvenaud.
\newblock Neural ordinary differential equations.
\newblock {\em Advances in Neural Information Processing Systems}, 31, 2018.

\bibitem{devlin2018bert}
Jacob Devlin, Ming-Wei Chang, Kenton Lee, and Kristina Toutanova.
\newblock Bert: Pre-training of deep bidirectional transformers for language
  understanding.
\newblock {\em arXiv preprint arXiv:1810.04805}, 2018.

\bibitem{duan2022multi}
Jiali Duan, Liqun Chen, Son Tran, Jinyu Yang, Yi~Xu, Belinda Zeng, and Trishul
  Chilimbi.
\newblock Multi-modal alignment using representation codebook.
\newblock In {\em Proceedings of the IEEE/CVF Conference on Computer Vision and
  Pattern Recognition}, pages 15651--15660, 2022.

\bibitem{gat2020removing}
Itai Gat, Idan Schwartz, Alexander Schwing, and Tamir Hazan.
\newblock Removing bias in multi-modal classifiers: Regularization by
  maximizing functional entropies.
\newblock {\em Advances in Neural Information Processing Systems},
  33:3197--3208, 2020.

\bibitem{goyal2017making}
Yash Goyal, Tejas Khot, Douglas Summers-Stay, Dhruv Batra, and Devi Parikh.
\newblock Making the v in vqa matter: Elevating the role of image understanding
  in visual question answering.
\newblock In {\em Proceedings of the IEEE Conference on Computer Vision and
  Pattern Recognition}, pages 6904--6913, 2017.

\bibitem{han2022multimodal}
Zongbo Han, Fan Yang, Junzhou Huang, Changqing Zhang, and Jianhua Yao.
\newblock Multimodal dynamics: Dynamical fusion for trustworthy multimodal
  classification.
\newblock In {\em Proceedings of the IEEE/CVF Conference on Computer Vision and
  Pattern Recognition}, pages 20707--20717, 2022.

\bibitem{han2021trusted}
Zongbo Han, Changqing Zhang, Huazhu Fu, and Joey~Tianyi Zhou.
\newblock Trusted multi-view classification.
\newblock {\em arXiv preprint arXiv:2102.02051}, 2021.

\bibitem{he2016deep}
Kaiming He, Xiangyu Zhang, Shaoqing Ren, and Jian Sun.
\newblock Deep residual learning for image recognition.
\newblock In {\em Proceedings of the IEEE Conference on Computer Vision and
  Pattern Recognition}, pages 770--778, 2016.

\bibitem{hong2020more}
Danfeng Hong, Lianru Gao, Naoto Yokoya, Jing Yao, Jocelyn Chanussot, Qian Du,
  and Bing Zhang.
\newblock More diverse means better: Multimodal deep learning meets
  remote-sensing imagery classification.
\newblock {\em IEEE Transactions on Geoscience and Remote Sensing},
  59(5):4340--4354, 2020.

\bibitem{hori2017attention}
Chiori Hori, Takaaki Hori, Teng-Yok Lee, Ziming Zhang, Bret Harsham, John~R
  Hershey, Tim~K Marks, and Kazuhiko Sumi.
\newblock Attention-based multimodal fusion for video description.
\newblock In {\em Proceedings of the IEEE International Conference on Computer
  Vision}, pages 4193--4202, 2017.

\bibitem{hou2019deep}
Ming Hou, Jiajia Tang, Jianhai Zhang, Wanzeng Kong, and Qibin Zhao.
\newblock Deep multimodal multilinear fusion with high-order polynomial
  pooling.
\newblock {\em Advances in Neural Information Processing Systems}, 32, 2019.

\bibitem{ioffe2015batch}
Sergey Ioffe and Christian Szegedy.
\newblock Batch normalization: Accelerating deep network training by reducing
  internal covariate shift.
\newblock In {\em International Conference on Machine Learning}, pages
  448--456. PMLR, 2015.

\bibitem{jayakumar2020multiplicative}
Siddhant~M. Jayakumar, Jacob Menick, Wojciech~M. Czarnecki, Jonathan Schwarz,
  Jack~W. Rae, Simon Osindero, Yee~Whye Teh, Tim Harley, and Razvan Pascanu.
\newblock Multiplicative interactions and where to find them.
\newblock In {\em International Conference on Learning Representations}, 2020.

\bibitem{karpathy2014large}
Andrej Karpathy, George Toderici, Sanketh Shetty, Thomas Leung, Rahul
  Sukthankar, and Li~Fei-Fei.
\newblock Large-scale video classification with convolutional neural networks.
\newblock In {\em Proceedings of the IEEE Conference on Computer Vision and
  Pattern Recognition}, pages 1725--1732, 2014.

\bibitem{kiros2015skip}
Ryan Kiros, Yukun Zhu, Russ~R Salakhutdinov, Richard Zemel, Raquel Urtasun,
  Antonio Torralba, and Sanja Fidler.
\newblock Skip-thought vectors.
\newblock {\em Advances in Neural Information Processing Systems}, 28, 2015.

\bibitem{li2021training}
Guohao Li, Matthias M{\"u}ller, Bernard Ghanem, and Vladlen Koltun.
\newblock Training graph neural networks with 1000 layers.
\newblock In {\em International Conference on Machine Learning}, pages
  6437--6449. PMLR, 2021.

\bibitem{liang2018multimodal}
Paul~Pu Liang, Ziyin Liu, Amir Zadeh, and Louis-Philippe Morency.
\newblock Multimodal language analysis with recurrent multistage fusion.
\newblock {\em arXiv preprint arXiv:1808.03920}, 2018.

\bibitem{liang2021multibench}
Paul~Pu Liang, Yiwei Lyu, Xiang Fan, Zetian Wu, Yun Cheng, Jason Wu, Leslie
  Chen, Peter Wu, Michelle~A Lee, Yuke Zhu, et~al.
\newblock Multibench: Multiscale benchmarks for multimodal representation
  learning.
\newblock {\em arXiv preprint arXiv:2107.07502}, 2021.

\bibitem{liang2019vrr}
Yuanzhi Liang, Yalong Bai, Wei Zhang, Xueming Qian, Li~Zhu, and Tao Mei.
\newblock Vrr-vg: Refocusing visually-relevant relationships.
\newblock In {\em Proceedings of the IEEE/CVF International Conference on
  Computer Vision}, pages 10403--10412, 2019.

\bibitem{liu2021contrastive}
Yunze Liu, Qingnan Fan, Shanghang Zhang, Hao Dong, Thomas Funkhouser, and
  Li~Yi.
\newblock Contrastive multimodal fusion with tupleinfonce.
\newblock In {\em Proceedings of the IEEE/CVF International Conference on
  Computer Vision}, pages 754--763, 2021.

\bibitem{liu2021swin}
Ze~Liu, Yutong Lin, Yue Cao, Han Hu, Yixuan Wei, Zheng Zhang, Stephen Lin, and
  Baining Guo.
\newblock Swin transformer: Hierarchical vision transformer using shifted
  windows.
\newblock In {\em Proceedings of the IEEE/CVF International Conference on
  Computer Vision}, pages 10012--10022, 2021.

\bibitem{liu2021group}
Ze~Liu, Zheng Zhang, Yue Cao, Han Hu, and Xin Tong.
\newblock Group-free 3d object detection via transformers.
\newblock In {\em Proceedings of the IEEE/CVF International Conference on
  Computer Vision}, pages 2949--2958, 2021.

\bibitem{liu2018efficient}
Zhun Liu, Ying Shen, Varun~Bharadhwaj Lakshminarasimhan, Paul~Pu Liang, Amir
  Zadeh, and Louis-Philippe Morency.
\newblock Efficient low-rank multimodal fusion with modality-specific factors.
\newblock {\em arXiv preprint arXiv:1806.00064}, 2018.

\bibitem{lu2021implicit}
Cheng Lu, Jianfei Chen, Chongxuan Li, Qiuhao Wang, and Jun Zhu.
\newblock Implicit normalizing flows.
\newblock {\em arXiv preprint arXiv:2103.09527}, 2021.

\bibitem{mroueh2015deep}
Youssef Mroueh, Etienne Marcheret, and Vaibhava Goel.
\newblock Deep multimodal learning for audio-visual speech recognition.
\newblock In {\em 2015 IEEE International Conference on Acoustics, Speech and
  Signal Processing (ICASSP)}, pages 2130--2134. IEEE, 2015.

\bibitem{nagrani2021attention}
Arsha Nagrani, Shan Yang, Anurag Arnab, Aren Jansen, Cordelia Schmid, and Chen
  Sun.
\newblock Attention bottlenecks for multimodal fusion.
\newblock {\em Advances in Neural Information Processing Systems},
  34:14200--14213, 2021.

\bibitem{natarajan2012multimodal}
Pradeep Natarajan, Shuang Wu, Shiv Vitaladevuni, Xiaodan Zhuang, Stavros
  Tsakalidis, Unsang Park, Rohit Prasad, and Premkumar Natarajan.
\newblock Multimodal feature fusion for robust event detection in web videos.
\newblock In {\em 2012 IEEE Conference on Computer Vision and Pattern
  Recognition}, pages 1298--1305. IEEE, 2012.

\bibitem{Nefian2002DynamicBN}
Ara~V. Nefian, Luhong Liang, Xiaobo Pi, Xiaoxing Liu, and Kevin~P. Murphy.
\newblock Dynamic bayesian networks for audio-visual speech recognition.
\newblock {\em EURASIP Journal on Advances in Signal Processing}, 2002:1--15,
  2002.

\bibitem{ngiam2011multimodal}
Jiquan Ngiam, Aditya Khosla, Mingyu Kim, Juhan Nam, Honglak Lee, and Andrew~Y
  Ng.
\newblock Multimodal deep learning.
\newblock In {\em International Conference on Machine Learning}, 2011.

\bibitem{ortega2019multimodal}
Juan~DS Ortega, Mohammed Senoussaoui, Eric Granger, Marco Pedersoli, Patrick
  Cardinal, and Alessandro~L Koerich.
\newblock Multimodal fusion with deep neural networks for audio-video emotion
  recognition.
\newblock {\em arXiv preprint arXiv:1907.03196}, 2019.

\bibitem{pan2020x}
Yingwei Pan, Ting Yao, Yehao Li, and Tao Mei.
\newblock X-linear attention networks for image captioning.
\newblock In {\em Proceedings of the IEEE/CVF Conference on Computer Vision and
  Pattern Recognition}, pages 10971--10980, 2020.

\bibitem{pennington2014glove}
Jeffrey Pennington, Richard Socher, and Christopher~D Manning.
\newblock Glove: Global vectors for word representation.
\newblock In {\em Proceedings of the 2014 Conference on Empirical Methods in
  Natural Language Processing (EMNLP)}, pages 1532--1543, 2014.

\bibitem{perez2019mfas}
Juan-Manuel P{\'e}rez-R{\'u}a, Valentin Vielzeuf, St{\'e}phane Pateux, Moez
  Baccouche, and Fr{\'e}d{\'e}ric Jurie.
\newblock Mfas: Multimodal fusion architecture search.
\newblock In {\em Proceedings of the IEEE/CVF Conference on Computer Vision and
  Pattern Recognition}, pages 6966--6975, 2019.

\bibitem{pham2019found}
Hai Pham, Paul~Pu Liang, Thomas Manzini, Louis-Philippe Morency, and
  Barnab{\'a}s P{\'o}czos.
\newblock Found in translation: Learning robust joint representations by cyclic
  translations between modalities.
\newblock In {\em Proceedings of the AAAI Conference on Artificial
  Intelligence}, volume~33, pages 6892--6899, 2019.

\bibitem{pineda1987generalization}
Fernando Pineda.
\newblock Generalization of back propagation to recurrent and higher order
  neural networks.
\newblock In {\em Neural Information Processing Systems}, 1987.

\bibitem{pokle2022deep}
Ashwini Pokle, Zhengyang Geng, and Zico Kolter.
\newblock Deep equilibrium approaches to diffusion models.
\newblock {\em arXiv preprint arXiv:2210.12867}, 2022.

\bibitem{qi2020imvotenet}
Charles~R Qi, Xinlei Chen, Or~Litany, and Leonidas~J Guibas.
\newblock Imvotenet: Boosting 3d object detection in point clouds with image
  votes.
\newblock In {\em Proceedings of the IEEE/CVF Conference on Computer Vision and
  Pattern Recognition}, pages 4404--4413, 2020.

\bibitem{qi2019deep}
Charles~R Qi, Or~Litany, Kaiming He, and Leonidas~J Guibas.
\newblock Deep hough voting for 3d object detection in point clouds.
\newblock In {\em Proceedings of the IEEE/CVF International Conference on
  Computer Vision}, pages 9277--9286, 2019.

\bibitem{ramachandram2017deep}
Dhanesh Ramachandram and Graham~W Taylor.
\newblock Deep multimodal learning: A survey on recent advances and trends.
\newblock {\em IEEE Signal Processing Magazine}, 34(6):96--108, 2017.

\bibitem{sankaran2021multimodal}
Sethuraman Sankaran, David Yang, and Ser-Nam Lim.
\newblock Multimodal fusion refiner networks.
\newblock {\em arXiv preprint arXiv:2104.03435}, 2021.

\bibitem{simonyan2014two}
Karen Simonyan and Andrew Zisserman.
\newblock Two-stream convolutional networks for action recognition in videos.
\newblock {\em Advances in Neural Information Processing Systems}, 27, 2014.

\bibitem{song2015sun}
Shuran Song, Samuel~P Lichtenberg, and Jianxiong Xiao.
\newblock Sun rgb-d: A rgb-d scene understanding benchmark suite.
\newblock In {\em Proceedings of the IEEE Conference on Computer Vision and
  Pattern Recognition}, pages 567--576, 2015.

\bibitem{sun2020learning}
Zhongkai Sun, Prathusha Sarma, William Sethares, and Yingyu Liang.
\newblock Learning relationships between text, audio, and video via deep
  canonical correlation for multimodal language analysis.
\newblock In {\em Proceedings of the AAAI Conference on Artificial
  Intelligence}, volume~34, pages 8992--8999, 2020.

\bibitem{teed2020raft}
Zachary Teed and Jia Deng.
\newblock Raft: Recurrent all-pairs field transforms for optical flow.
\newblock In {\em European Conference on Computer Vision}, pages 402--419.
  Springer, 2020.

\bibitem{tsai2019multimodal}
Yao-Hung~Hubert Tsai, Shaojie Bai, Paul~Pu Liang, J~Zico Kolter, Louis-Philippe
  Morency, and Ruslan Salakhutdinov.
\newblock Multimodal transformer for unaligned multimodal language sequences.
\newblock In {\em Proceedings of the conference. Association for Computational
  Linguistics. Meeting}, volume 2019, page 6558. NIH Public Access, 2019.

\bibitem{tsai2018learning}
Yao-Hung~Hubert Tsai, Paul~Pu Liang, Amir Zadeh, Louis-Philippe Morency, and
  Ruslan Salakhutdinov.
\newblock Learning factorized multimodal representations.
\newblock {\em arXiv preprint arXiv:1806.06176}, 2018.

\bibitem{van2016better}
Mark~A Van De~Wiel, Tonje~G Lien, Wina Verlaat, Wessel~N van Wieringen, and
  Saskia~M Wilting.
\newblock Better prediction by use of co-data: adaptive group-regularized ridge
  regression.
\newblock {\em Statistics in Medicine}, 35(3):368--381, 2016.

\bibitem{vaswani2017attention}
Ashish Vaswani, Noam Shazeer, Niki Parmar, Jakob Uszkoreit, Llion Jones,
  Aidan~N Gomez, {\L}ukasz Kaiser, and Illia Polosukhin.
\newblock Attention is all you need.
\newblock {\em Advances in Neural Information Processing Systems}, 30, 2017.

\bibitem{vielzeuf2018centralnet}
Valentin Vielzeuf, Alexis Lechervy, St{\'e}phane Pateux, and Fr{\'e}d{\'e}ric
  Jurie.
\newblock Centralnet: a multilayer approach for multimodal fusion.
\newblock In {\em Proceedings of the European Conference on Computer Vision
  (ECCV) Workshops}, pages 0--0, 2018.

\bibitem{wang2021mogonet}
Tongxin Wang, Wei Shao, Zhi Huang, Haixu Tang, Jie Zhang, Zhengming Ding, and
  Kun Huang.
\newblock Mogonet integrates multi-omics data using graph convolutional
  networks allowing patient classification and biomarker identification.
\newblock {\em Nature Communications}, 12(1):1--13, 2021.

\bibitem{wang2020deep}
Yikai Wang, Wenbing Huang, Fuchun Sun, Tingyang Xu, Yu~Rong, and Junzhou Huang.
\newblock Deep multimodal fusion by channel exchanging.
\newblock {\em Advances in Neural Information Processing Systems},
  33:4835--4845, 2020.

\bibitem{wu2018group}
Yuxin Wu and Kaiming He.
\newblock Group normalization.
\newblock In {\em Proceedings of the European Conference on Computer Vision
  (ECCV)}, pages 3--19, 2018.

\bibitem{Xu2018TexttoClipVR}
Huijuan Xu, Kun He, Leonid Sigal, Stan Sclaroff, and Kate Saenko.
\newblock Text-to-clip video retrieval with early fusion and re-captioning.
\newblock {\em arXiv preprint arXiv:1804.05113}, 2018.

\bibitem{xue2022dynamic}
Zihui Xue and Radu Marculescu.
\newblock Dynamic multimodal fusion.
\newblock {\em arXiv preprint arXiv:2204.00102}, 2022.

\bibitem{yang2020cm}
Kaicheng Yang, Hua Xu, and Kai Gao.
\newblock Cm-bert: Cross-modal bert for text-audio sentiment analysis.
\newblock In {\em Proceedings of the 28th ACM International Conference on
  Multimedia}, pages 521--528, 2020.

\bibitem{ye2012robust}
Guangnan Ye, Dong Liu, I-Hong Jhuo, and Shih-Fu Chang.
\newblock Robust late fusion with rank minimization.
\newblock In {\em 2012 IEEE Conference on Computer Vision and Pattern
  Recognition}, pages 3021--3028. IEEE, 2012.

\bibitem{yu2019deep}
Zhou Yu, Jun Yu, Yuhao Cui, Dacheng Tao, and Qi~Tian.
\newblock Deep modular co-attention networks for visual question answering.
\newblock In {\em Proceedings of the IEEE/CVF Conference on Computer Vision and
  Pattern Recognition}, pages 6281--6290, 2019.

\bibitem{zadeh2018memory}
Amir Zadeh, Paul~Pu Liang, Navonil Mazumder, Soujanya Poria, Erik Cambria, and
  Louis-Philippe Morency.
\newblock Memory fusion network for multi-view sequential learning.
\newblock In {\em Proceedings of the AAAI Conference on Artificial
  Intelligence}, volume~32, 2018.

\bibitem{zadeh2018multi}
Amir Zadeh, Paul~Pu Liang, Soujanya Poria, Prateek Vij, Erik Cambria, and
  Louis-Philippe Morency.
\newblock Multi-attention recurrent network for human communication
  comprehension.
\newblock In {\em Proceedings of the AAAI Conference on Artificial
  Intelligence}, volume~32, 2018.

\bibitem{zadeh2016mosi}
Amir Zadeh, Rowan Zellers, Eli Pincus, and Louis-Philippe Morency.
\newblock Mosi: multimodal corpus of sentiment intensity and subjectivity
  analysis in online opinion videos.
\newblock {\em arXiv preprint arXiv:1606.06259}, 2016.

\end{thebibliography}
}


\newpage
\appendix
\onecolumn
\section{Proof for Backpropagation of DEQ Fusion}\label{appendix:proof-backward}
\begin{proofidx}{\ref{thm:loss-x-grad}}
Our proof is similar to \cite{bai2019deep}. We know $\mathbf{z}_i^*=f_{\theta_i}(\mathbf{z}_i^*;\mathbf{x}_i)$ from \cref{eq:fixed-point}, we can first differentiate two sides implicitly with respect to $\mathbf{x}_i$:
\begin{equation}\label{eq:dzi*-dxi}
\begin{aligned}
    \frac{\mathrm{d}\mathbf{z}_i^*}{\mathrm{d}\mathbf{x}_i}&=\frac{\mathrm{d}f_{\theta_i}(\mathbf{z}_i^*;\mathbf{x}_i)}{\mathrm{d}\mathbf{x}_i}\\
    &=\frac{\partial f_{\theta_i}(\mathbf{z}_i^*;\mathbf{x}_i)}{\partial\mathbf{x}_i}+\frac{\partial f_{\theta_i}(\mathbf{z}_i^*;\mathbf{x}_i)}{\partial \mathbf{z}_i^*}\cdot\frac{\mathrm{d}\mathbf{z}_i^*}{\mathrm{d}\mathbf{x}_i}
\end{aligned}
\end{equation}
Rearranging \cref{eq:dzi*-dxi}, we obtain
\begin{equation}
    \left(\mathbf{I}-\frac{\partial f_{\theta_i}(\mathbf{z}_i^*;\mathbf{x}_i)}{\partial\mathbf{z}_i^*}\right)\frac{\mathrm{d}\mathbf{z}_i^*}{\mathrm{d}\mathbf{x}_i}=\frac{\partial f_{\theta_i}(\mathbf{z}_i^*;\mathbf{x}_i)}{\partial\mathbf{x}_i}.
\end{equation}
Differentiating \cref{eq:residual-g-theta} with respect to $\mathbf{z}_i^*$, we obtain the Jacobian
\begin{equation}\label{eq:jac-theta}
    {J_{g_{\theta_i}}}|_{\mathbf{z}_i^*}=-\left(\mathbf{I}-\frac{\partial f_{\theta_i}(\mathbf{z}_i^*;\mathbf{x}_i)}{\partial\mathbf{z}_i^*}\right)
\end{equation}
Therefore $\frac{\mathrm{d}\mathbf{z}_i^*}{\mathrm{d}\mathbf{x}_i}=\left(-{J_{g_{\theta_i}}^{-1}}|_{\mathbf{z}_i^*}\right)\cdot\frac{\partial f_{\theta_i}(\mathbf{z}_i^*;\mathbf{x}_i)}{\partial\mathbf{x}_i}$.

Similarly, we have $\mathbf{z}_{\mathrm{fuse}}^*=f_{\mathrm{fuse}}(\mathbf{z}_{\mathrm{fuse}}^*;\mathbf{x}_{\mathrm{fuse}})$ from \cref{eq:fixed-point}. Differentiating both sides with respect to $\mathbf{z}_i^*$:
\begin{equation}\label{eq:dzfuse*-dzi*}
\begin{aligned}
    \frac{\mathrm{d}\mathbf{z}_{\mathrm{fuse}}^*}{\mathrm{d}\mathbf{z}_i^*}&=\frac{\mathrm{d}f_{\mathrm{fuse}}(\mathbf{z}_{\mathrm{fuse}}^*;\mathbf{x}_{\mathrm{fuse}})}{\mathrm{d}\mathbf{z}_i^*}\\
    &=\frac{\partial f_{\mathrm{fuse}}(\mathbf{z}_{\mathrm{fuse}}^*;\mathbf{x}_{\mathrm{fuse}})}{\partial\mathbf{z}_i^*}+\frac{\partial f_{\mathrm{fuse}}(\mathbf{z}_{\mathrm{fuse}}^*;\mathbf{x}_{\mathrm{fuse}})}{\partial \mathbf{z}_{\mathrm{fuse}}^*}\cdot\frac{\mathrm{d}\mathbf{z}_{\mathrm{fuse}}^*}{\mathrm{d}\mathbf{z}_i^*}
\end{aligned}
\end{equation}
Rearranging \cref{eq:dzfuse*-dzi*}, we have
\begin{equation}
    \left(\mathbf{I}-\frac{\partial f_{\mathrm{fuse}}(\mathbf{z}_{\mathrm{fuse}}^*;\mathbf{x}_{\mathrm{fuse}})}{\partial\mathbf{z}_{\mathrm{fuse}}^*}\right)\frac{\mathrm{d}\mathbf{z}_{\mathrm{fuse}}^*}{\mathrm{d}\mathbf{z}_i^*}=\frac{\partial f_{\mathrm{fuse}}(\mathbf{z}_{\mathrm{fuse}}^*;\mathbf{x}_{\mathrm{fuse}})}{\partial\mathbf{z}_{\mathrm{fuse}}^*}.
\end{equation}
Similar to computation in \cref{eq:jac-theta}, we have:
\begin{equation}
    \left.{J_{g_{\mathrm{fuse}}}}\right|_{\mathbf{z}_{\mathrm{fuse}}^*}=-\left(\mathbf{I}-\frac{\partial f_{\mathrm{fuse}}(\mathbf{z}_{\mathrm{fuse}}^*;\mathbf{x}_{\mathrm{fuse}})}{\partial\mathbf{z}_{\mathrm{fuse}}^*}\right).
\end{equation}
Thus $\frac{\mathrm{d}\mathbf{z}_{\mathrm{fuse}}^*}{\mathrm{d}\mathbf{z}_i^*}=\left(-\left.{J_{g_{\mathrm{fuse}}}^{-1}}\right|_{\mathbf{z}_{\mathrm{fuse}}^*}\right)\cdot\frac{\partial f_{\mathrm{fuse}}(\mathbf{z}_{\mathrm{fuse}}^*;\mathbf{x}_{\mathrm{fuse}})}{\partial\mathbf{z}_{\mathrm{fuse}}^*}$.

Finally, we can differentiate loss $\ell$ with respect to $\mathbf{x}_i$:
\begin{equation}
\begin{aligned}
    \frac{\partial\ell}{\partial\mathbf{x}_i}&=\frac{\partial\ell}{\partial\mathbf{z}_{\mathrm{fuse}}^*}\cdot\frac{\mathrm{d}\mathbf{z}_{\mathrm{fuse}}^*}{\mathrm{d}\mathbf{z}_i^*}\cdot\frac{\mathrm{d}\mathbf{z}_i^*}{\mathrm{d}\mathbf{x}_i}\\
    &=\frac{\partial\ell}{\partial\mathbf{z}_{\mathrm{fuse}}^{*}}\cdot\left(\left.{-J_{g_{\mathrm{fuse}}}^{-1}}\right|_{\mathbf{z}_{\mathrm{fuse}}^{*}}\right)\cdot\frac{\partial f_{\mathrm{fuse}}\left(\mathbf{z}_{\mathrm{fuse}}^{*};\mathbf{x}_{\mathrm{fuse}}\right)}{\partial\mathbf{z}_i^*}\cdot\left({-J_{g_{\theta_i}}^{-1}}|_{\mathbf{z}_{i}^{*}}\right)\cdot\frac{\partial f_{\theta_i}\left(\mathbf{z}_{i}^{*};\mathbf{x}_{i}\right)}{\partial\mathbf{x}_i} 
\end{aligned}
\end{equation}
\end{proofidx}

\section{Experimental Setup}\label{sec:setup}
We conduct the experiments on NVIDIA Tesla V100 GPUs and use Anderson acceleration \cite{anderson1965iterative} as the default fixed point solver for all our experiments.

\textbf{BRCA.}\quad We experiment based on the current state-of-the-art approach \cite{han2022multimodal} by replacing the original concatenation fusion with our DEQ fusion. Following \cite{han2022multimodal}, the learning rate is set to 0.0001 and decays at the rate of 0.2 every 500 steps. As the dataset is relatively small, we additionally leverage dropout in fusion layer and early stopping to prevent overfitting. Jacobian regularization loss with a loss weight of 20 is employed to stabilize training. We report the mean and standard deviation of the experimental results over 10 runs.

\textbf{MM-IMDB.}\quad Our implementation and experiments on MM-IMDB are based on MultiBench \cite{liang2021multibench}. We follow the data split and feature extraction methods presented in \cite{arevalo2017gated} for data preprocessing. Jacobian regularization loss with a loss weight of 0.1 is exploited. To further stabilize training, we additionally set a smaller learning rate of 0.0001 for our DEQ fusion module, and 0.001 for all other weights.

\textbf{CMU-MOSI.}\quad We conduct the experiments with the state-of-the-art CM-BERT \cite{yang2020cm} by replacing the original simple addition fusion strategy with our DEQ fusion. We follow \cite{bai2021stabilizing} and use Jacobian regularization loss with a loss weight of 0.01 to stabilize DEQ training.

\textbf{SUN RGB-D.}\quad We conduct the experiments based on ImVoteNet \cite{qi2020imvotenet}. We use the public train-test split (5,285 vs 5,050). We follow the hyperparameter settings and training details in the officially released codebase\footnote{\url{https://github.com/facebookresearch/imvotenet}} except that we trained the models on 4 GPUs with a batch size of 32 for 140 epochs for fast convergence. 

\textbf{VQA-v2.}\quad Our experiments are based on Mutan \cite{ben2017mutan} and MCAN \cite{yu2019deep}. All methods are trained on the train set (444k samples) and evaluated on the validation set (214k samples). Our Mutan\footnote{\url{https://github.com/Cadene/vqa.pytorch}} and MCAN\footnote{\url{https://github.com/MILVLG/mcan-vqa}} results are  reproduced based on their official codebases respectively. For a fair comparison, we apply the bottom-up-attention visual features for all experiments and only use the VQA-v2 training set (disabled VisualGenome and VQA-v2 val set) for model training. Our reproduced Mutan baseline has better performance than the other reproduced version in \cite{liang2019vrr} (63.73\% vs. 62.84\% in overall accuracy) under the same settings. For MCAN, we select its ``Large'' model setting as our baseline.   

\section{Additional Ablation Studies}\label{sec:add-ablation}
We additionally conduct ablation studies on MM-IMDB and CMU-MOSI, the results are shown in \cref{table:add-ablation}. The same experimental setup as demonstrated in \cref{sec:setup} is used. Note that if $f_{\mathrm{fuse}}$ is not used, $G(\cdot)$ is automatically disabled (denoted as ``-''). The conclusions are similar to the one made in \cref{sec:ablation}, except that we do not observe the performance drop with our additional $f_{\theta}$ and $f_{\mathrm{fuse}}$ (first row and second row). A potential reason is that BRCA is a relatively small dataset, and thus can be easily overfitted with more weights. Nonetheless, all empirical results demonstrate that DEQ fusion with all proposed components leads to the most superior results.

\begin{table}[t]
\centering
\caption{Ablation experiments on MM-IMDB and CMU-MOSI. ``-'' indicates not applicable.} 
\label{table:add-ablation}
\vskip 0.1in
\centering
\small
\begin{tabular}{cccc|cc|ccccc}
\toprule
& & & & \multicolumn{2}{c|}{MM-IMDB} & \multicolumn{5}{c}{CMU-MOSI} \\
$f_{\theta}$ & $f_{\mathrm{fuse}}$ & DEQ & $G(\cdot)$ & MicroF1 & MacroF1 & Acc-7 & Acc-2 & F1 & MAE & Corr \\
\midrule
 & & & - & 58.76 & 49.63 & 43.3 & 83.3 & 83.2 & 0.755 & 0.786 \\
 $\checkmark$ & $\checkmark$ & & $\checkmark$ & 60.73 & 52.64 & 43.0 & 83.6 & 83.6 & 0.757 & 0.787 \\
$\checkmark$ & & $\checkmark$ & - & 59.80 & 49.27 & 43.7 & 84.8 & 84.9 & 0.741 & 0.782 \\
 & $\checkmark$ & $\checkmark$ & $\checkmark$ & 60.76 & 53.09 & 45.3 & 84.4 & 84.3 & 0.747 & 0.782 \\
$\checkmark$ & $\checkmark$ & $\checkmark$ &  & 60.83 & 52.67 & 43.8 & 83.1 & 83.1 & 0.751 & 0.789 \\
$\checkmark$ & $\checkmark$ & $\checkmark$ & $\checkmark$ & \textbf{61.52} & \textbf{53.38} & \textbf{46.1} & \textbf{85.4} & \textbf{85.4} & \textbf{0.737} & \textbf{0.797} \\
\bottomrule
\end{tabular}
\vspace{-4mm}
\end{table}

\begin{table}[!t]
\centering
\caption{Convergence of DEQ Fusion. The values indicate the relative difference norm computed at a given solver step.} 
\label{table:add-convergence}
\vskip 0.1in
\centering
\small
\begin{tabular}{lccccc}
\toprule
Dataset & step 1 & step 10 & step 20 & step 40 & step 100 \\
\midrule
BRCA & 7.06e-1 & 3.38e-2 & 8.80e-3 & 5.18e-3 & 1.29e-3 \\
MM-IMDB & 2.86e-1 & 9.17e-4 & 7.65e-5 & 8.87e-6 & 2.17e-6 \\
CMU-MOSI & 3.09e-2 & 4.16e-7 & 6.94e-8 & 5.66e-8 & 5.66e-8 \\
\bottomrule
\end{tabular}
\vspace{-4mm}
\end{table}

In addition to the convergence ablation study on BRCA, we further examine the convergence of DEQ fusion on MM-IMDB and CMU-MOSI. The results are in \cref{table:add-convergence}. DEQ fusion successfully converges on all three benchmarks, whereas the convergence rate on MM-IMDB and CMU-MOSI is considerably faster than on BRCA.

\end{document}